\pdfoutput=1

\documentclass[11pt]{article}

\usepackage{EMNLP2022}
\usepackage{times}
\usepackage{latexsym}
\usepackage{pdflscape}
\usepackage{rotfloat}
\usepackage[T1]{fontenc}

\usepackage[utf8]{inputenc}

\usepackage{microtype}

\usepackage{inconsolata}
\usepackage{amsmath}
\usepackage{amsthm}
\usepackage[cmintegrals]{newtxmath}
\usepackage{booktabs}
\usepackage{graphicx}
\newcommand{\eg}{\textit{e}.\textit{g}.}

\usepackage{multirow}
\usepackage{color}
\usepackage{stfloats}

\title{Video Question Answering: Datasets, Algorithms and Challenges}

\author{Yaoyao Zhong\textsuperscript{\rm 1,3}\thanks{\quad The first three authors contribute equally to this work.}\ ,
Junbin Xiao\textsuperscript{\rm 1,2}$^*$,
    Wei Ji\textsuperscript{\rm 1,2}$^*$,\\
    \bf{Yicong Li\textsuperscript{\rm 1},
    Weihong Deng\textsuperscript{\rm 3}\thanks{\quad Corresponding authors.}\ ,
    Tat-Seng Chua\textsuperscript{\rm 1,2}} \\
    \textsuperscript{\rm 1}National University of Singapore, Singapore\quad
    \textsuperscript{\rm 2}Sea-NExT Joint Lab, Singapore\\
	\textsuperscript{\rm 3}Beijing University of Posts and Telecommunications, Beijing, China\\
	\texttt{\small{\{zhongyaoyao,whdeng\}@bupt.edu.cn,}}
	\texttt{\small{junbin@comp.nus.edu.sg}}, \\ 
	\texttt{\small{liyicong@u.nus.edu}},
	\texttt{\small{\{jiwei,dcscts\}@nus.edu.sg}}
}
\begin{document}
\maketitle

\begin{abstract}
This survey aims to organize the recent advances in video question answering (VideoQA) and point towards future directions. 
We firstly categorize the datasets into: 1) normal VideoQA, multi-modal VideoQA and knowledge-based VideoQA, according to the modalities invoked in the question-answer pairs, and 2) factoid VideoQA and inference VideoQA, according to the technical challenges in comprehending the questions and deriving the correct answers. 
We then summarize the VideoQA techniques, including those mainly designed for Factoid QA (such as the early spatio-temporal attention-based methods and the recent Transformer-based ones) and those targeted at explicit relation and logic inference (such as neural modular networks, neural symbolic methods, and graph-structured methods). Aside from the backbone techniques, we also delve into specific models and derive some common and useful insights either for video modeling, question answering, or for cross-modal correspondence learning. 
Finally, we present the research trends of studying beyond factoid VideoQA to inference VideoQA, as well as towards the robustness and interpretability. 
Additionally, we maintain a repository, \url{https://github.com/VRU-NExT/VideoQA}, to keep trace of the latest VideoQA papers, datasets, and their open-source implementations if available. 
With these efforts, we strongly hope this survey could shed light on the follow-up VideoQA research.

\end{abstract}

\section{Introduction}
Recent years have witnessed a flourish of research in vision-language understanding~\citep{xu2016msr,chen2017sca,antol2015vqa,chen2018temporally,jang2017tgif}, of which, video Question Answering (VideoQA) is one of the most prominent, given its promise to develop interactive AI to communicate with the dynamic visual world via natural languages. 
Despite the popularity, VideoQA remains one of the greatest challenges, because it demands the models to comprehensively understand the videos to correctly answer questions. The questions involve not only the recognition of visual objects, actions, activities and events, but also the inference of their semantic, spatial, temporal, and causal relationships~\citep{xu2017video,jang2017tgif,shang2019annotating,shang2021video,yang2021deconfounded,xiao2021next,xiao2021video}.

To tackle the challenges, techniques such as spatio-temporal attention \citep{jang2017tgif}, motion-appearance memory \citep{gao2018motion}, and spatio-temporal or hierarchical graph models \cite{cherian2022,xiao2021video} have been proposed and demonstrated their effectiveness on different VideoQA datasets. However, we find that the datasets, the defined challenges, and the corresponding algorithms are varied and a bit messy. There is a lack of a meaningful survey to categorize the datasets and to organize the technique developed, which seriously impedes the research. 

\begin{sidewaysfigure}[p]
\centering
\scalebox{0.52}{
\includegraphics[height=0.475\textheight]{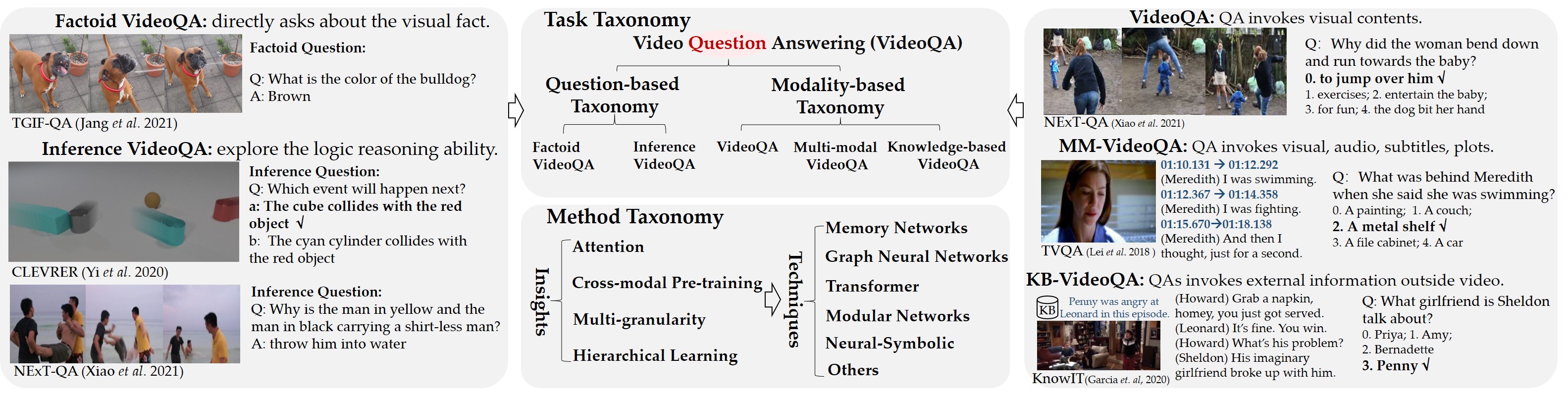}
}
\caption{Illustration of the taxonomy. The taxonomy covers not only Factoid and Inference VideoQA in terms of understanding level, but also VideoQA, Multi-modal VideoQA, and Knowledge-based VideoQA in terms of the multi-modal information invoked in QAs to better analyze the challenges and help to uncover the future focus. }
\label{fig:introduction}
\end{sidewaysfigure}

Although a handful of recent works~\citep{sun2021video,khurana2021video,patel2021recent} have tried to review VideoQA, they mostly follow an old-to-new fashion to summarize the literature and lack an effective taxonomy to classify them. In terms of the contents, these works focus merely on factoid questions and neglect the inference questions (see Fig.~\ref{fig:introduction} for the  difference). Furthermore, lots of recent new techniques (\eg, pre-training and Transformer) are missing.

This paper thus gives a more comprehensive and meaningful survey to VideoQA, in the hope of learning from the past and shaping the future. 
Our contributions are as follows.
(1) We provide a clear taxonomy to VideoQA. We can either classify existing VideoQA tasks into Factoid VideoQA and Inference VideoQA according to the fundamental challenges embodied in QAs, or classify them into normal VideoQA, Multi-modal VideoQA, and Knowledge-based VideoQA according to the multi-modal information invoked in the QAs. 
(2) We categorize existing VideoQA techniques as Memory, Transformer, Graph, Neural Modular Network, and Neural-Symbolic method. Along with the techniques, some meaningful insights are also included: attention modeling, cross-modal pre-training, hierarchical learning, multi-granular ensemble, and progressive reasoning.
(3) We analyze existing methods from the perspective of the challenges encountered in the various VideoQA tasks and provide our prospects for future research.

\section{VideoQA Task and Datasets}
\subsection{Problem Formulation}
VideoQA is a task to predict the correct answer $a^*$ based on a question $q$ and a video $V$. There are mainly two types of tasks in VideoQA: multi-choice QA and open-ended QA. 

For \textbf{multi-choice} QA, the models are presented with several candidate answers $\mathcal{A}_{mc}$ for each question and are required to pick the correct one $a^*=\mathcal{F}\left( a|q,\mathcal{V},\mathcal{A}_{mc} \right)$. 
For \textbf{open-ended} QA, the problem can be classification (the most popular), generation (word-by-word) and regression (for counting) depending on the specific datasets. Specifically, open-ended QA is popularly set as a multi-class classification problem which requires the models to classify a video-question pair into a pre-defined global answer set $\mathcal{A}_{oe}$: $a^*=\mathcal{F}\left( a|q,\mathcal{V} \right)$ where $a\in\mathcal{A}_{oe}$. Open-ended QA can also be formulated as a generation problem, which might have more practical use and receiving increasing attention. Usually the answer is denoted as ${a}=(a_1,a_2,...,a_t,...,a_M)$ of length $M$, where $a_t$ is the $t$-th word; and the model is required to predict the next word $a_t$ in the vocabulary set $\mathcal{W}$: 
$a_t^* = \mathcal{F}\left( a_t|q,\mathcal{V}, (a_1,a_2,...,a_{t-1})\right)$,
where $a_t\in\mathcal{W}$. For the counting task, which is defined as an open-ended question about counting the number of repetitions of an action~\cite{jang2017tgif}, it is formulated as an regression problem, requiring the model to compute an integer-valued answer to be close to the ground truth. 

Compared with open-ended QA, multi-choice QA is typically defined to study beyond factoid QA to inference QA \cite{xiao2021next,wu2021star}, as it dispenses with the generation and evaluation of natural languages.

\subsection{Evaluation Metrics}

\textbf{Accuracy.} For multi-choice QA and open-ended QA (classification), accuracy is defined based on the entire testing question set $\mathcal{Q}$, given by: 
\begin{equation}
\setlength{\abovedisplayskip}{3pt}
acc=\frac{1}{\left| \mathcal{Q} \right|}\sum_{q\in \mathcal{Q}}{\mathbf{I}\left[ a^*=a \right]},
\setlength{\belowdisplayskip}{3pt}
\end{equation}
where $\mathcal{Q}$ represents the number of QA pairs, and $\mathbf{I}\left[ \cdot \right]$ is an indicator function (1 only if $a^* = a$ and 0 otherwise). Similarly, for open-ended QA (word-by-word generation) ~\cite{zhao2017videob,zhao2018open}, accuracy is defined as:
\begin{equation}
\setlength{\abovedisplayskip}{3pt}
    acc=\frac{1}{\left| \mathcal{Q} \right|}\sum_{q\in \mathcal{Q}}{\frac{1}{M}\sum_{i=1}^{L}{\mathbf{I}\left[ a_{i}^{*}=a_i \right]}},
\setlength{\belowdisplayskip}{3pt}
\end{equation}
where $L$ denotes the length of the shorter answer. 

\textbf{WUPS.} The WUPS is the soft measure of accuracy by taking into account word synonyms. It is based on the WUP score~\cite{DBLP:conf/acl/WuP94} to evaluate the quality of the generated answer~\cite{zhao2017videob,zhao2018open,xiao2021next}. The WUP measures word similarity based on WordNet~\cite{fellbaum1998wordnet}. WUPS score with the threshold $\gamma$ is defined as,
\begin{equation}\begin{small}
\begin{gathered}
WUPS=\frac{1}{\left| \mathcal{Q} \right|}\sum_{q\in \mathcal{Q}}{\min \{ \prod_{a_i\in A}{\underset{a_{j}^{*}\in A^*}{\max}WUP_{\gamma}( a_i,a_{j}^{*})}, }\hfill \\
\qquad\qquad\qquad\qquad\qquad\quad {\prod_{a_{i}^{*}\in A^*}{\underset{a_j\in A}{\max}WUP_{\gamma}( a_{i}^{*},a_j )} \}},
\end{gathered}\end{small}
\end{equation}where WUP score is given by,
	\begin{equation}\begin{scriptsize}WUP_{\gamma}( a_i,a_{j}^{*}) = 
    \begin{cases}
			WUP( a_i,a_{j}^{*} )&		WUP( a_i,a_{j}^{*} ) \geqslant \gamma\\
			0.1WUP( a_i,a_{j}^{*} )&		WUP( a_i,a_{j}^{*} ) <\gamma\\
	\end{cases}.
	\end{scriptsize}\end{equation}
where the parameter $\gamma$ is dataset-specific.
	
\textbf{Mean $\mathcal{L}_2$ loss.} For the repetition count task~\cite{jang2017tgif}, the mean $\mathcal{L}_2$ loss is defined based on the entire testing question set $\mathcal{Q}$: 
\begin{equation}
\setlength{\belowdisplayskip}{3pt}
\mathcal{L}_2 =\frac{1}{\left| \mathcal{Q} \right|}\sum_{q\in \mathcal{Q}}{\left( a^*-a \right)^2},
\setlength{\belowdisplayskip}{3pt}
\end{equation}
in which a and a* are predicted and ground-truth numbers respectively.

The evaluation metrics mainly serve for different task settings, while there are also some novel and diagnostic ones~\cite{gandhi2022measuring,li2022representation,castro2022wild} that may be helpful for robustness and interpretation of VideoQA models.

\begin{sidewaysfigure}[p]
\centering
\scalebox{1}{
\includegraphics[width=1.0\linewidth]{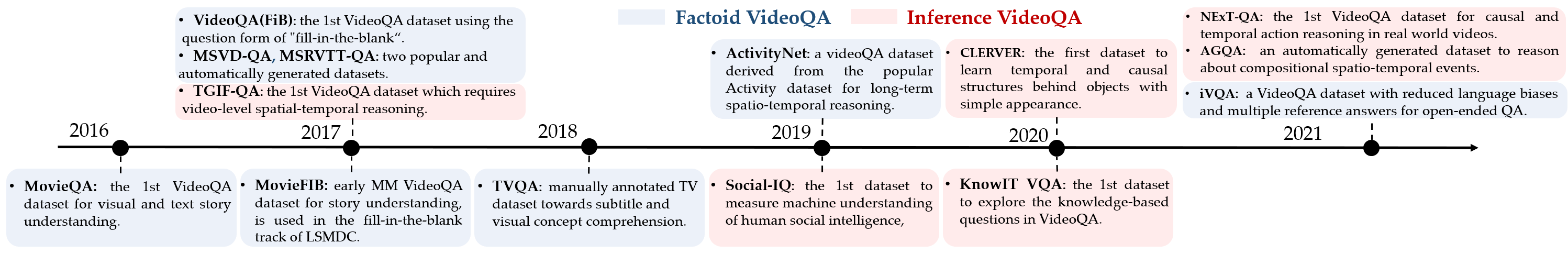}
}
\caption{Timeline of established VideoQA datesets. The items above the timeline show the normal VideoQA datasets. The Multi-modal VideoQA and Knowledge-based VideoQA are listed below the timeline. Blue and red colors represent datasets focused on Factoid VideoQA and Inference VideoQA.}
\label{fig:history}
\end{sidewaysfigure}

\subsection{Datasets}
VideoQA can be understood from different perspectives, since the aim is to gain multi-view and multi-grained understanding of videos under the guidance of specific questions.

\setlength{\tabcolsep}{20pt}
\begin{table*}[t]
		\renewcommand\arraystretch{1.2}
		\center
		\caption{VideoQA datasets in the literature.}
        \label{table:table_datasets}
        \scalebox{0.6}{
        \begin{tabular}{@{}l|c|c@{}}
        \hline\hline
        & Factoid VideoQA & Inference VideoQA \\ \hline
        VideoQA & \begin{tabular}[c]{@{}c@{}} 
        
        VideoQA(FiB)~\citep{zhu2017uncovering}, VideoQA~\citep{zeng2017leveraging}, 
        \\ MSVD-QA~\citep{xu2017video}, 
        MSRVTT-QA~\citep{xu2017video}, 
        \\ YouTube2Text-QA~\citep{zhao2017videoa}, 
        MarioQA~\citep{mun2017marioqa}, 
        \\  EgoVQA~\citep{fan2019egovqa},
        ActivityNet-QA~\citep{yu2019activitynet}, \\ iVQA~\citep{yang2021just},
        ASRL-QA~\citep{sadhu2021video}, \\ Charades-SRL-QA~\citep{sadhu2021video},  FIBER~\citep{castro2022fiber}, \\ WildQA~\citep{castro2022wild} \end{tabular} 
        & \begin{tabular}[c]{@{}c@{}}TGIF-QA~\citep{jang2017tgif},  
        SVQA~\citep{song2018explore}, 
        \\ V2C-QA~\citep{fang2020video2commonsense}, CLEVRER~\citep{yi2020clevrer}, 
        \\ SUTD-TrafficQA~\citep{xu2021sutd}, 
        \\AGQA~\small{\citep{grunde2021agqa}}, 
        \\ AGQA 2.0~\citep{gandhi2022measuring}, 
        VQuAD~\citep{gupta2022vquad}, 
        \\ STAR~\citep{wu2021star},
        \\ NExT-QA~\citep{xiao2021next},Causal-VidQA~\citep{li2022representation}

        \end{tabular} \\ \hline
        MM VideoQA 
        & \begin{tabular}[c]{@{}c@{}}MovieQA~\citep{tapaswi2016movieqa}, MovieFIB~\citep{maharaj2017dataset}, 
        \\ PororoQA~\citep{kim2017deepstory}, TVQA~\citep{lei2018tvqa}, 
        \\ TVQA+~\citep{lei2020tvqa+}, 
        LifeQA~\citep{castro2020lifeqa}, \\ How2QA~\citep{li2020hero}, Env-QA~\citep{gao2021env}, 
        \\ Pano-AVQA~\citep{yun2021pano}, DramaQA~\citep{choi2021dramaqa},
        \\MUSIC-AVQA~\citep{li2022learning}, AVQA~\cite{yang2022avqa}
        \end{tabular}                                      
        & Social-IQ~\citep{zadeh2019social} \\ \hline
        KB VideoQA & / 
        & \begin{tabular}[c]{@{}c@{}} PsTuts-VQA~\citep{DBLP:conf/ijcai/ZhaoKXJ20}, 
        KnowIT VQA~\citep{garcia2020knowit}, 
        \\ KnowIT-X VQA~\citep{wu2021transferring},  \\ NEWSKVQA~\citep{gupta2022newskvqa} \end{tabular}  \\ 
        \hline
        \end{tabular}
	    }
\end{table*}

\textbf{Modality-based Taxonomy.} According to the data modality invoked in the questions and answers, VideoQA can be classified into normal VideoQA, multi-modal VideoQA (MM VideoQA), and knowledge VideoQA (KB VideoQA). Normal VideoQA only invokes visual resources to understand the question and to derive the correct answer. It emphasizes visual understanding of the video elements and reasoning of their relations. Usually, the videos are short and are typically user-generated on social platforms. Different from normal VideoQA, MM VideoQA often involves other resources aside from visual contents, such as subtitles/transcripts and text plots of movies~\citep{tapaswi2016movieqa} and TV shows~\citep{lei2018tvqa}. MM VideoQA mainly challenges multi-modal information fusion and long video story understanding. Finally, KB VideoQA~\citep{garcia2020knowit} demands external knowledge distillation from explicit knowledge bases or commonsense reasoning~\citep{fang2020video2commonsense}. Different from MM VideoQA, KB VideoQA provides a global knowledge base for the whole dataset, instead of giving paired ``knowledge'' for each question. For better understanding of the three kinds of VideoQA, we show typical examples in Figure~\ref{fig:introduction} (right).

\textbf{Question-based Taxonomy.} According to the type of question (or the challenges posted in the questions), VideoQA can be classified into factoid VideoQA and inference VideoQA. A factoid question directly asks about the visual fact, such as the location (\texttt{where is}), objects/attributes (\texttt{who/what (color) is}), and invokes little relations to understand the questions and infer the correct answers. Factoid QA emphasizes the holistic understanding of the questions and the recognition of the visual elements. In contrast, inference VideoQA aims to explore the logic and knowledge reasoning ability in dynamic scenarios. It features various relationships between the visual facts. Though rich in relation types, VideoQA emphasizes temporal (\texttt{before/after}) and causal (\texttt{why/How/what if}) relationships that feature temporal dynamics, as emphasized by recent works \citep{zadeh2019social,yi2020clevrer,xiao2021next,li2022representation}.

\textbf{Datasets Analysis.} The timeline of some established VideoQA datasets is shown in Figure~\ref{fig:history}. We categorize all the datasets according to our defined taxonomy in Table~\ref{table:table_datasets} and their details are listed in Table~\ref{table:table_details_datasets} (see Appendix). VideoQA and MM VideoQA almost appear simultaneously, and have been studied separately by the community. Despite the unique challenges of MM VideoQA in reasoning on multiple modalities~\citep{kim2020modality}, algorithms targeting VideoQA and MM VideoQA share similar spirits. Modality-based taxonomy stems from research preference for video domains. While question-based taxonomy is affected more by the methodological considerations, since the recently proposed Inference VideoQA brings new technical challenges, which is driving artificial intelligence towards new heights, not just limited to learning the correlations in data. 

\subsection{Main Framework}
As shown in Figure~\ref{fig:framework}, a common framework comprises four parts: video encoder, question encoder, cross-modal interaction, and answer decoder. The video encoder often encodes raw videos by jointly extracting frame appearance and clip motion features. Recent works also show that object-level visual and semantic features (\eg, category and attribute labels) are important. These features are usually extracted with pre-trained 2D or 3D neural networks, as summarized in Table~\ref{table:table_factoid}. Question encoder extracts token-level representation, such as GloVe and BERT features~\citep{devlin2018bert}. Then, the sequential data of vision and language can be further processed by sequential models (\eg, RNN, CNN, and Transformer) for the convenience of cross-modal interaction, which will be detailed further. For multi-choice QA, the answer decoder can be a 1-way classifier to select the correct answer from the provided multiple choices. For open-ended QA, it can be either an n-way classifier to select an answer from a pre-defined global answer set, or a language generator to generate an answer word by word. The video and language encoders can be pre-trained or more recently end-to-end fine-tuned~\citep{lei2021less}. 

\subsection{Challenges and Meaningful Insights}
\textbf{Unique Challenges.} Compared with ImageQA~\citep{lu2016hierarchical,anderson2018bottom}, VideoQA is much more challenging because of the spatio-temporal nature of videos \cite{xiao2020visual,xiao2021next}. Thus, a simple extension of existing ImageQA techniques to answer queries of videos will lead to sub-optimal results. Compared with other video tasks, question-answering requires a comprehensive understanding of videos in different aspects and granularity, such as from fine-grained to coarse-grained in both temporal and spatial domains, and from factoid questions to inference questions. To tackle the challenges, a lot of research efforts have been developed on cross-modal interaction, which aims to gain understanding of videos under the guidance of questions. We summarize some common and meaningful insights as follows:

\begin{figure}[t!]
	\center
	\includegraphics[width=1\linewidth]{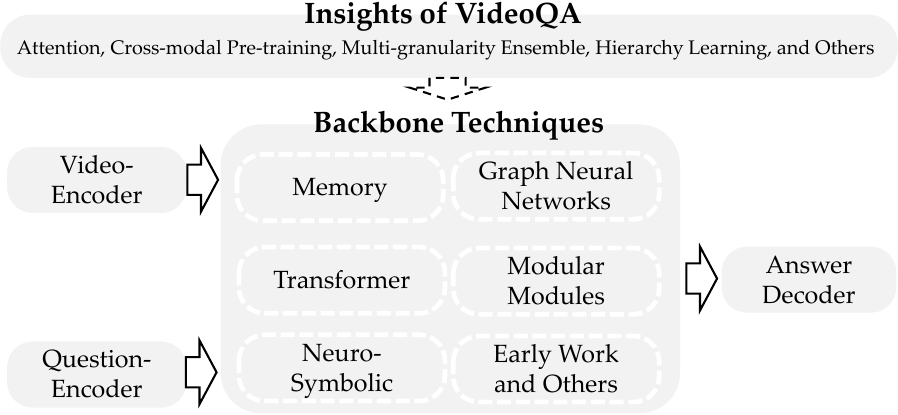}
	\caption{A common solution framework for VideoQA. It includes: a video encoder, a question encoder, a cross-modal interaction, and an answer decoder. Some common insights are also involved in the model design.}
	\label{fig:framework}
\end{figure}

\textbf{Attention.} Attention is a human-inspired mechanism that locates the important part of the input and selectively focuses on useful information. In VideoQA, to attend to a specific part of videos in both spatial and temporal dimensions, \textit{temporal attention} and \textit{spatial attention} are widely used. \textit{Self-attention} has a good ability to model long-range dependencies, and can be used in intra-modal modeling, such as temporal information in the video and global dependencies of questions. \textit{Co-attention} (\textit{Cross-modal attention}) can attend to both relevant and critical multi-modal information, such as the question-guided video representation and video-guided question representation.   

\textbf{Cross-modal Pre-training and Fine-tuning.} With the development of unified network architectures (\eg, Transformer~\citep{vaswani2017attention}) that can well handle visual and linguistic data, cross-modal pre-training can make full use of the semantic information from noisy but web-scale vision-text data~\citep{radford2021learning}. The learned model can be transferred to downstream vision-language tasks by fine-tuning on small-scale manually annotated datasets with strong supervision.
Currently, the research in this paradigm lies in four aspects: large-scale data collection, proxy task definition, Transformer-style model design, and downstream adaption. We recommend the readers to read the latest survey~\citep{chen2022vlp} for details.


\textbf{Multi-Granularity Ensemble.} Questions are diverse and unconstrained, and may demand video information of different granularities for answers~\citep{xiao2021video}. To gain rich information and answer the varied questions, the multi-granularity ensemble is essential. Specifically, the multi-granularity ensemble exists in both the text domain and the vision domain of both spatial and temporal dimensions. In the text domain, word-, phrase- and sentence-level feature representations are coordinated to achieve both fine- and coarse-grained information modeling. In the vision domain, region-, trajectory-, frame- and clip-level feature representations can complement each other to achieve comprehensive video understanding. 


\textbf{Hierarchical Learning.} Considering that the video elements and their textual correspondences in QA pairs are in different abstraction levels, hierarchical learning aims to organize multi-modal representation from low-level to high-level, and from local to global~\citep{le2020hierarchical,dang2021hierarchical,xiao2021video}. Specifically, linguistic concepts are analyzed from word to sentence. Similarly, video elements are processed from objects to actions, activities, and global events. Compared with the multi-granularity ensemble, hierarchical learning processes the multi-granular information progressively. It gradually reasons and aggregates the low-level, local visual information into the high-level, global video representation. Thus, hierarchical learning can better reflect the structure and relationship of video elements and accomplish question answering hierarchically. 

\textbf{Others.} Aside from the above, multi-step reasoning~\citep{wang2021dualvgr,mao2022dynamic} and causal discovery~\citep{li2022invariant} also demonstrate the effectiveness. Most importantly, these insights are not mutually exclusive; they can be coordinated in a single model for good performance.

\section{Algorithms}
\subsection{Methods}
\quad\textbf{Early Attention-based Works.} 
\citep{zeng2017leveraging} try to directly apply element-wise multiplication to fuse the global video and question representations for answer prediction. Additionally, it demonstrates the advantage of a simple temporal attention. Attention is also explored in more complex scenarios in conjunction with various other ideas, such as multi-granularity ensemble~\citep{xu2017video} and hierarchical learning~\citep{zhao2017videoa}. In particular,
\citep{jang2017tgif} propose a dual-LSTM based approach with spatial and temporal attention mechanisms, which can focus better on critical frames in a video and critical regions in a frame. \citep{xu2017video} refine attention over both frame-level and clip-level visual features, conditioned with both the coarse-grained question feature and fine-grained word feature. \citep{zhao2017videoa} propose hierarchical dual-level attention networks (DLAN) to learn the question-aware video representations with word-level and question-level attention based on appearance and motion.

Despite the ability to attend to video frames and clips, these works rely on RNN for history information modeling, which has later been shown to be weak in capturing long-term dependency.

\textbf{Memory Networks.} Memory networks can cache sequential inputs in memory slots and explicitly utilize even far early information. Memory especially receives attention in \textit{long video story understanding}, such as movies and TV-Shows. Because the QAs in these VideoQA tasks not only involve the understanding of visual contents, but also the long stories they convey. 

\citep{tapaswi2016movieqa} first incorporate and modify the memory network~\citep{sukhbaatar2015end} into VideoQA, to store video and subtitle features in the memory bank. To enable memory read and write operations with high capacity and flexibility, \citep{Na_2017_ICCV} design a memory network with multiple convolution layers. Considering dual-modal information in the movie story, \citep{kim2019progressive} introduce a progressive attention mechanism to progressively prune out irrelevant temporal parts in the memory bank for each modality, and adaptively integrate outputs of each memory. 

Memory has also been explored in \textit{normal VideoQA}. \citep{gao2018motion} propose a two-stream framework (CoMem) to deal with motion and appearance information with a co-memory attention module, introducing multi-level contextual information and producing dynamic fact ensembles for diverse questions. Considering that CoMem synchronizes the attentions detected by appearance and motion features, it could thus generate incorrect attention, \citep{fan2019heterogeneous} further introduce a heterogeneous external memory module (HME) with attentional read and write operations to integrate the motion and appearance features and learn the spatio-temporal attention simultaneously. 


\textbf{Transformer.} 
Transformer~\citep{vaswani2017attention} has a good ability to model long-term relationships and has demonstrated promising performance for modeling multi-modal vision-language tasks such as VideoQA, with pre-training on large-scale datasets~\citep{zhu2020actbert}. Motivated by the success of Transformer, \citep{li2019beyond} first introduce the architecture of Transformer \textit{without pre-training} to VideoQA (PSAC), which consists of two positional self-attention blocks to replace LSTM, and a video-question co-attention block to simultaneously attend both visual and textual information. \citep{yang2020bert} and \citep{urooj2020mmft} incorporate the pre-trained \textit{language-based} Transformer (BERT)~\citep{devlin2018bert} to movie and story understanding, which requires more modeling on languages like subtitles and dialogues. Both works process each of the input modalities such as video and subtitles, with question and candidate answer, respectively, and lately fuse several streams for the final answer.  

More recently, \citep{lei2021less} apply the \textit{image-text pre-trained Transformer} for cross-modal pre-training and fine-tune it for downstream video-text tasks, such as VideoQA. \citep{yang2021just} train a VideoQA model, based on a large-scale dataset, with 69M video-question-answer triplets, using contrastive learning between a multi-modal video-question Transformer and an answer Transformer. This \textit{video-text pre-trained Transformer} can be further fine-tuned on other downstream VideoQA tasks, which shows the benefits of task-specific pre-training for the target VideoQA task. Furthermore, \citep{zellers2021merlot} train a cross-modal Transformer (MERLOT) in a label-free, self-supervised manner, based on 180M video segments with image frames and words. Similar to MERLOT, VIOLET~\citep{fu2021violet} is another end-to-end \textit{video-text pre-trained} Transformer model but with more advanced video encoder and proxy tasks. 

While the aforementioned Transformer-style models have demonstrated strong performances on popular Factoid VideoQA datasets (refer to our analysis in Sec.~\ref{sec:result}), recent works \cite{buch2022revisiting,xiao2022video} reveal that their performance are weak in answering questions that emphasize visual relation reasoning, especially the temporal and causal relations which feature video dynamics. Furthermore, their demands on large-scale video data for pre-training and the lack of explanability largely prevent their popularity. Such weaknesses call for more future efforts in developing foundation models for fine-grained video reasoning, and simultaneously, with less computation resources and better interpretability. 

\textbf{Graph Neural Networks.}
Graph-structured techniques~\citep{kipf2017semi,zhang2022fine} are recently more favoured for improving the reasoning ability of VideoQA models, especially when Inference VideoQA draws attention to the community~\citep{jang2017tgif,xiao2021next}. HGA~\citep{jiang2020reasoning}, and more recent works, B2A~\citep{park2021bridge} and DualVGR~\citep{wang2021dualvgr} build the graphs based on coarse-grained video segments. Yet, they incorporate both intra- and inter-modal relationship learning and achieve good performances. To gain object-level information, \citep{huang2020location} build the graph (LGCN) based on objects represented by their appearance and location features. They model the interaction between objects related to questions with GNN~\citep{kipf2017semi}.
 
Considering that the video elements are hierarchical in semantic space,  \citep{liu2021hair}, \citep{peng2021progressive} and \citep{xiao2021video} incorporate hierarchical learning into graph networks. Specifically, \citep{liu2021hair} propose a graph memory mechanism (HAIR),  to perform relational vision-semantic reasoning from object level to frame level; \citep{peng2021progressive} concatenate different-level graphs, that is, object-level, frame-level, and clip-level, progressively to learn the visual relations (PGAT). \citep{xiao2021video} propose a hierarchical conditional graph model (HQGA) to weave together visual facts from low-level entities to higher-level video elements through graph aggregation and pooling, to enable vision-text matching at multi-granularity levels. To leverage the semantics of the 3D scene, \citep{cherian2022} transfer the video frames to a 2.5D (pseudo-3D) scene graph and then split it into static and dynamic sub-graphs, allowing the pruning of redundant detections.

With a good ability for information communication, graph architectures have shown promising results on inference VideoQA. Nonetheless, the emphasis and difficulty lie in how to skillfully design the graph structure for video representation.

\textbf{Modular Networks.} \citep{le2020hierarchical} find that most VideoQA models design tailor-made network architectures. They point out such hand-crafted architectures are inflexible in dealing with varied data modality, video length and question types. Therefore, they design a reusable neural unit - Conditional Relation Network (CRN), which captures the relations of input features given the global context and encapsulates them hierarchically to form networks. Such a constituted architecture has shown better generalization ability and flexibility in handling different types of questions. Following similar design philosophy, \citep{dang2021hierarchical} and \citep{xiao2021video} design the spatio-temporal graph and conditional graph respectively as neural building blocks. The neural building blocks are hierarchically stacked to achieve good reasoning performances. While the above works aim for repeating a single module for videoQA. Recently, \cite{Qian2022DynamicSM} design multiple modules tailored for compositional video question-answering \cite{grunde2021agqa}, and has also demonstrated success. Overall, modular networks are of improved flexibility and transparency. Nonetheless, they either lack explicit logic for reasoning \cite{le2020hierarchical,dang2021hierarchical,xiao2021video}, or can only handle questions that can be parsed into pre-defined subtasks of limited scope.


\textbf{Neural-Symbolic.} \citep{yi2020clevrer} point out two essential elements for causal reasoning in VideoQA are object-centric video representation that is aware of the temporal and causal relations between the objects and events, and a dynamics model that is able to predict the object dynamics under unobserved or counterfactual scenarios. Motivated by the neural-symbolic method in ImageQA~\citep{yi2018neural}, \citep{yi2020clevrer} propose the NS-DR model, which extracts object-level representation with a video parser, turns a question into a functional program, extracts and predicts the dynamic scene of the video with a dynamics predictor, and runs the program on the dynamic scene to obtain an answer. NS-DR aims to combine neural nets for pattern recognition and dynamics prediction, and symbolic logic for causal reasoning. It achieves significant gain on the explanatory, predictive, and counterfactual questions on the synthetic object dataset~\citep{yi2020clevrer}. \citep{chen2021grounding} and~\citep{ding2021dynamic} promote further progress. 

Despite the good reasoning ability of Neural-Symbolic methods on synthetic datasets, they are currently hard to be applied in unconstrained video with open-form natural questions. 

\setlength{\tabcolsep}{14pt}
\begin{table*}[ht]
	\renewcommand\arraystretch{1.2}
	\center
	\caption{Performance on Factoid VideoQA tasks. (Att: Attention, MG: Multi-Granularity, HL: Hierarchical Learning, CM-PF: Cross-modal Pre-training and Fine-tuning, Mem: Memory, GNN: Graph Neural Networks, MN: Modular Networks, TF: Transformer. RN: ResNet at frame-level, RX(3D): 3D ResNeXt at clip-level, RoI: Region-of-interest features from Faster R-CNN, GV: GloVe, BT: BERT, VG: Visual Genome \protect\citep{krishna2017visual}
	, YT-T: Youtube-Temporal-180M \protect\citep{zellers2021merlot}
	, Web: WebVid2M \protect\citep{bain2021frozen}
	, CC: Conceptual Captions-3M \protect\citep{sharma2018conceptual}
	. ViT \protect\citep{dosovitskiy2020image} 
	and VSwin \protect\citep{liu2021video}
	are Transformer-style visual encoders. Attention is found in all methods, but we omit it for those methods that do not emphasize attention.) }
	\label{table:table_factoid}
	\scalebox{0.595}{
		\begin{tabular}{l|l|ll|c|c|c|c}
			\hline\hline
			\multirow{2}{*}{Methods}  & \multirow{2}{*}{Techniques \& Insights} & \multicolumn{2}{c|}{Encoder} & \multirow{2}{*}{\begin{tabular}[c]{@{}c@{}} Pre-training\\ Dataset\end{tabular}} & \multirow{2}{*}{\begin{tabular}[c]{@{}c@{}} TGIF-QA\\ (Frame-QA)\end{tabular}}& \multirow{2}{*}{\begin{tabular}[c]{@{}c@{}} MSVD\\ -QA\end{tabular}}& \multirow{2}{*}{\begin{tabular}[c]{@{}c@{}}MSRVTT\\ -QA\end{tabular}} \\ \cline{3-4} 
			& & \multicolumn{1}{c|}{Video}     & Text  &  &  &  &  \\ \hline
			STVQA\citep{jang2019video} & Att & \multicolumn{1}{c|}{RN, Flow}                            & GV  & / & 52.0 & / & / \\
			
			PSAC\citep{li2019beyond} & Att & \multicolumn{1}{c|}{RN}   & GV  
			& / & 55.7 & / & /\\
			
			QueST\citep{jiang2020divide} & Att & \multicolumn{1}{c|}{RN, C3D} & GV  & / & 59.7 & 36.1 & 34.6                         \\
			\hline
			
			CoMem\citep{gao2018motion}        & Mem                       & \multicolumn{1}{c|}{RN, Flow}                  & GV  & /                           & 51.5   & /                        & /                          \\
			HME\citep{fan2019heterogeneous}   & Mem                            & \multicolumn{1}{c|}{RN, VGG, C3D}                  & GV  & /            & 53.8 & 33.7 & 33.0\\
			
			\hline
			LGCN\citep{huang2020location}     & GNN                    & \multicolumn{1}{c|}{RN, RoI}                  & GV  & /                                        & 56.3                     & 34.3                     & /                          \\
			
			HGA\citep{jiang2020reasoning}     & GNN                      & \multicolumn{1}{c|}{RN, VGG, C3D}                  & GV  & /                                              & 55.1                                & 34.7                     & 35.5                       \\
			B2A\citep{park2021bridge}         & GNN, MG                     & \multicolumn{1}{c|}{RN, RX(3D)}               & GV  & /                                         & 57.5           & 37.2                     & 36.9                       \\
			HAIR\citep{liu2021hair}           & GNN, Mem, HL                      & \multicolumn{1}{c|}{RoI}                      & GV  & /                            & 60.2                        & 37.5                     & 36.9                       \\

			MASN\citep{seo2021attend}         & GNN                  & \multicolumn{1}{c|}{RN, I3D, RoI} & GV                          &                   & 59.5              & 38.0                    & 35.2             \\
			DualVGR\citep{wang2021dualvgr}    & GNN                 & \multicolumn{1}{c|}{RN, RX(3D)}  & GV  & /                                                                    & /             & 39.0                     & 35.5                       \\

			PGAT\citep{peng2021progressive}   & GNN, MG, HL                     & \multicolumn{1}{c|}{RN, RX(3D), RoI} & GV  & /                         & 61.1         & 39.0                     & 38.1                       \\

			\hline
			HCRN\citep{le2020hierarchical}    & MN, HL                        & \multicolumn{1}{c|}{RN, RX(3D)}               & GV  & /                                & 55.9                     & 36.1                     & 35.6                       \\
			HOSTR\citep{dang2021hierarchical} & MN, GNN, HL                         & \multicolumn{1}{c|}{RN, RX(3D), RoI}               & GV  & /                                     & 58.2                            & 39.4            & 35.9                       \\
			
			HQGA\citep{xiao2021video}         & MN, GNN, HL, MG                  & \multicolumn{1}{c|}{RN, RX(3D), RoI} & BT   & /                                            & 61.3           & 41.2                     & 38.6             \\
			
			\hline
			MHN\citep{peng2022multilevel}         & TF, HL, MG                  & \multicolumn{1}{c|}{RN, RX(3D)} & GV   & /                & 58.1           & 40.4                     & 38.6              \\
			VGT\citep{xiao2022video}         & TF, GNN                  & \multicolumn{1}{c|}{RN, RoI} & BT   & /    & 61.6           & /                     & 39.7              \\
			
			ClipBERT\citep{lei2021less}      & TF, CM-PF                         & \multicolumn{1}{c|}{RN (E2E)}                            & BT & VG\&COCO                                   & 60.3                                 & /                        & 37.4                       \\
			CoMVT\citep{seo2021look}         & TF, CM-PF                         & \multicolumn{1}{c|}{S3D}                               & BT   & HowTo100M                               & /             & 42.6                     & 39.5                       \\
			VQA-T\citep{yang2021just}         & TF, CM-PF                         & \multicolumn{1}{c|}{S3D}                               & BT   & H2VQA69M               & /                                    & 46.3                     & 41.5                       \\
			SiaSRea\citep{yu2021learning}   & TF, GNN, CM-PF                         & \multicolumn{1}{c|}{RN (E2E)}                            & BT & VG\&COCO                & 60.2             & 45.5                        & 41.6                       \\
			MERLOT\citep{zellers2021merlot}   & TF, CM-PF                        & \multicolumn{1}{c|}{ViT(E2E)}                            & BT & YT-T \& CC                            & \textbf{69.5}            & /                        & 43.1                       \\
			VIOLET\citep{fu2021violet}        & TF, CM-PF                        & \multicolumn{1}{c|}{VSwin (E2E)}                            & BT & Web\&YT-T\&CC                                     & 68.9                               & \textbf{47.9}            & \textbf{43.9}              \\ \hline
		\end{tabular}
	}
\end{table*}

\textbf{Others.} There are also \textit{flexibly designed networks} to address specific problems. For example, \citep{kim2020modality} propose a framework that first detects a specific temporal moment from moments of interest candidates for temporally-aligned video and subtitle using pre-defined sliding windows, and then fuses information based on the localized moment using intra-modal and cross-modal attention mechanisms. Due to their focuses on specific purposes, the question remains on whether these networks can be generalized to other VideoQA tasks. 


Studies are also conducted in terms of \textit{input information}. 
\citep{falcon2020data} explore several \textit{data augmentation} techniques to prevent overfitting with only small-scale datasets. 
\citep{kim2021video} point out existing works suffer from significant computational complexity and insufficient representation capability and they introduce VideoQA features obtained from \textit{coded video bit-stream} to address the problem. To overcome spurious visual-linguistic correlations, \citep{li2022invariant,li2022equivariant} explore robust and trustworthy grounding framework from causal theory, which is promising to enhance the SOTA models' accuracy and trustability. 

While the above efforts focus on a better video representation for question answering, a handful of works \citep{xue2018better,hong2019learning} also pay attention to the language side by reserving the syntactic structure \cite{FeiMatchStruICML22,FeiLatDSCWWW21} of the questions, and also shows advantages.

\subsection{Performance Analysis} 
\label{sec:result}
We analyze the advanced methods for Factoid VideoQA in Table~\ref{table:table_factoid} and Inference VideoQA in Table~\ref{table:table_tc} based on the results reported on popular VideoQA benchmarks. Apart from normal VideoQA, advanced methods for MM VideoQA and KB VideoQA are also summarized in Table~\ref{table:table_mkn}. 

\setlength{\tabcolsep}{3pt}
\begin{table}[t]
	\renewcommand\arraystretch{1.2}
	\center
	\caption{Performance on Inference VideoQA tasks. For the counting (Cnt) task in TGIF-QA, value of mean square error (MSE) is reported for evaluation.} 
	\label{table:table_tc}
	\scalebox{0.6}{
		\begin{tabular}{l|l|cc|ccc}
			\hline\hline
			\multirow{2}{*}{Methods} & \multirow{2}{*}{Techniques \& Insights} &\multicolumn{2}{c|}{NExT-QA} & \multicolumn{3}{c}{TGIF-QA} \\ \cline{3-7}
			&  &  Val. & Test & Act  &  Tran. &  Cnt \\ \hline
			STVQA\small{\citep{jang2017tgif}} & Att & 47.9 & 47.6 & 62.9 & 69.4 & 4.22 \\
			CoMem\small{\citep{gao2018motion}}  & Mem & 48.0 & 48.5 & 68.2 & 74.3 & 4.10 \\
			HME\citep{fan2019heterogeneous} & Mem  & 48.7      & 49.2 & 73.9 & 77.8 & 4.02 \\
			HCRN\small{\citep{le2020hierarchical}}  & MN, HL    &  48.2 & 48.9 & 75.0 & 81.4 & 3.82 \\
			HGA\small{\citep{jiang2020reasoning}}   & GNN, HL    &  49.7     & 50.0 & 75.4 & 81.0 & 4.09 \\
			MASN\small{\citep{seo2021attend}} & GNN & / &/ & 84.4 & 87.4 & 3.75 \\
			MHN\small{\citep{peng2022multilevel}} & TF, HL, MG & / & / & 83.5 & 90.2 & {\bf3.57} \\
			IGV\small{\citep{li2022invariant}}       & GNN, Causal &  51.0     & 51.3    & / & / & /       \\
			HQGA\small{\citep{xiao2021video}}       & MN, GNN, HL, MG & 51.4     & 51.8  & 76.9 & 85.6 & /  \\
			P3D-G\small{\citep{cherian2022}}       & GNN, TF, HL    &  53.4     & / & / & / & /          \\
			ATP\small{\citep{buch2022revisiting}} & TF  &  54.3     & /  & / & / & /         \\
			VGT\small{\citep{xiao2022video}} & TF,GNN  &  \textbf{55.0}     & \textbf{53.7}   & {\bf95.0} & {\bf97.6} & /         \\
			\hline
			ClipBERT\small{\citep{lei2021less}} & TF, CM-PF & / & / & 82.8 & 87.8 & / \\ 
			SiasRea\small{\citep{yu2021learning}} & TF, CM-PF, GNN & / & / & 79.7 & 85.3 & / \\
			MERLOT\small{\citep{zellers2021merlot}} & TF, CM-PF & / & / & 94.0 & 96.2 & / \\
			VIOLET\small{\citep{fu2021violet}} &TF, CM-PF & / & / & 92.5 & 95.7 & / \\
			\hline
 			Human & / &  \textbf{88.4} & / & / & / & / \\  
            \hline
		\end{tabular}
	}
\end{table}

Table~\ref{table:table_factoid} reveals that the cross-modal pre-trained Transformer-style models can achieve superior performance for factiod QA than others. By focusing on methods without pre-training, graph-structured techniques are the most popular and have also shown great potential. It would be interesting to explore cross-modal pre-training of graph models for VideoQA. Besides, hierarchical learning and fine-grained object features usually help to improve performances. In addition to the datasets given in Table~\ref{table:table_factoid}, the recent iVQA \cite{yang2021just} dataset has also received increasing attention, and we believe it could be a more effective dataset towards open-ended VideoQA for its high quality.

Inference VideoQA is a nascent task that challenges mainly visual relation reasoning of video information. It also receives increasing attention. Graph-structured techniques, causal discovery, and hierarchical learning have shown promising performance (see Table~\ref{table:table_tc}). Notably, we find that cross-modal pre-training and fine-tuning not only achieves good performance on factoid VideoQA, but also significantly improves the results on inference VideoQA. Particularly, the accuracies of reasoning tasks on TGIF-QA reach unprecedentedly high. This dataset is likely not challenging enough and has serious language bias as revealed by recent studies \cite{peng2021progressive,piergiovanni2022video,xiao2022video}. In contrast, NExT-QA is much more challenging; it emphasizes causal and temporal relation reasoning between multiple objects in real-world videos. Table~\ref{table:table_tc} shows that SOTA methods still struggle on NExT-QA. As such, NExT-QA could be a more effective benchmark for visual reasoning of realistic video contents under natural language instructions. Additionally, NExT-QA also contains open-ended QA task that provide ample challenge for existing research.  

\setlength{\tabcolsep}{5pt}
\begin{table}[t!]
	\renewcommand\arraystretch{1.2}
	\center
	\caption{Performance on MM VideoQA and KB VideoQA tasks. For TVQA, we report results on test-public data split. (ts: Timestamp Annotation.) }
	\label{table:table_mkn}
	\scalebox{0.595}{
\begin{tabular}{l|l|cc|c|c}
\hline\hline
\multirow{2}{*}{Methods} & \multirow{2}{*}{\begin{tabular}[c]{@{}c@{}}Techniques \& \\ Insights \end{tabular}}& \multicolumn{2}{c|}{TVQA}           & \multirow{2}{*}{\small{TVQA+}} & \multirow{2}{*}{\begin{tabular}[c]{@{}c@{}}\small{KnowIT}\\ \small{VQA}\end{tabular}} \\ \cline{3-4} & & \multicolumn{1}{c|}{\small{w/o ts}} & \small{w/ ts} &                        &           \\ \hline

PAMN\small{\citep{kim2019progressive}} &  Mem & \multicolumn{1}{c|}{66.8}   & /     & /                      & /                           \\

STAGE\small{\citep{lei2020tvqa+}} & Att & \multicolumn{1}{c|}{\textbf{70.2}}   & /     & 74.8                   & / \\

HCRN\small{\citep{le2021hierarchical}} & MN, HL& \multicolumn{1}{c|}{66.1} & 71.3 & / & / \\ 

MSAN\small{\citep{kim2020modality}} & Att & \multicolumn{1}{c|}{/}      & 71.1  & / & /     \\

BERT-VQA\tiny{\citep{devlin2018bert}} & TF & \multicolumn{1}{c|}{/} & 73.6  & / & / \\

MMFT-BERT\small{\citep{urooj2020mmft}}  & TF & \multicolumn{1}{c|}{/} & 72.9  & / & /  \\

ROLL\small{\citep{garcia2020knowledge}} & TF & \multicolumn{1}{c|}{/}      & /     & 69.6 & 71.5                        \\
RHA\small{\citep{li2021relation}} & GNN, HL & \multicolumn{1}{c|}{/}      & /     & 73.4 & /                        \\
SPCR\small{\citep{kim2021self}} & Att & \multicolumn{1}{c|}{/} & 76.2  & 76.2 & / \\

V2T\small{\citep{engin2021hidden}} & TF & \multicolumn{1}{c|}{/} & / & /  & \textbf{78.1}                        \\
MERLOT\small{\citep{zellers2021merlot}} &   TF, CM-PF & \multicolumn{1}{c|}{/}      & \textbf{78.7}  & \textbf{80.9}                   & /                           \\\hline

Human &  / & \multicolumn{1}{c|}{\textbf{89.4}} & \textbf{91.5}  & \textbf{90.5} & /                           \\ \hline
\end{tabular}
	}
\end{table}

MM and KB VideoQA require models to locate and perform reasoning in all heterogeneous modalities for answering the question. Similar to normal VideoQA, MM VideoQA also benefits from advanced networks and large-scale datasets. However, it is worth noting that modality shifting ability is essential~\citep{kim2020modality,engin2021hidden}.

\section{Future Direction}


\textbf{From Recognition to Reasoning.}
Advanced neural network models excel at recognizing objects, attributes and even actions in visual data. Thus, answering the questions like "what is" is no longer the core of VideoQA. To enable more meaningful and in-depth human-machine interaction, it is urgent to study the casual and temporal relations between objects, actions, and events~\citep{xiao2021next}. Such problems feature \emph{video}-level understanding and demand inference ability for question answering. The focus on inference questions promotes research towards the core of human intelligence, which could be one of the ``north stars'' towards groundbreaking works~\citep{fei2022searching}.

\textbf{Knowledge VideoQA.} 
To answer the questions that are beyond the visual scene, it is of crucial importance to inject knowledge into the reasoning stage~\citep{jin2019video,garcia2020knowit,zhuang2020multichannel,wu2021transferring}. Knowledge incorporation can not only greatly extend the scope of questions that can be asked about videos, but also enable the exploration of more human-like inference. Because we humans are natural to answer questions that may involve commonsense~\citep{fang2020video2commonsense,chadha2020iperceive} or domain-specific knowledge~\citep{xu2021sutd,gao2021env}. Reasoning with knowledge and diagnosing the retrieved knowledge for a specific question will help to enhance the model's interpretability and trustability. It will also serve as important groundwork for the future multi-modal conversation systems~\citep{nie2019multimodal,li2022mmcoqa}. 

\textbf{Cross-modal Pre-training and Fine-tuning.}   
Cross-modal pre-trained representations \citep{zellers2021merlot,fu2021violet} have shown great benefit for VideoQA (see Table~\ref{table:table_factoid} and \ref{table:table_tc}). However, most models only demonstrate their good performance on VideoQA tasks that challenge the recognition or shallow description of the video contents. Also, it demands a lot of computation and other resources to handle large-scale video-text data. Therefore, how to pre-train vision-language models more efficiently and how to adapt them to reasoning type of VideoQA tasks deserve more attention. 

\textbf{Interpretability, Robustness and Generalization.}
Despite the strong power of the advanced pre-training models, it is still unknown how they work, to what extent they can generalize, when they will fail, and how to gain further technical improvement. Recent works towards interpretability and logical robustness~\citep{li2021adversarial,sheng2021human} have achieved initial success. \citep{gandhi2022measuring} design a benchmark to diagnose whether models can gain true understanding by examining compositional consistency. However, there is a still long way to go towards model interpretability, robustness and generalization. We believe this is of great significance towards practical multi-modal QA systems and multi-modal conversation systems.


\section{Conclusion}
This paper gives a quick overview to the broad aspect of video question answering. We mainly categorized the related datasets and techniques. Also, we discussed some meaningful insights and analyzed the performances of different techniques on different type of datasets. We finally concluded several promising future directions. With these efforts, we hope this survey can shed light and attract more research to VideoQA, and eventually, foster more efforts towards strong AI systems that can demonstrate their understanding of the dynamic visual world by making meaningful responses to our natural language instructions or queries.

\section*{Acknowledgements}
The research is supported by the Sea-NExT Joint Lab. The research is also supported by the National Natural Science Foundation of China (No.62236003 and No.62276030), and China Scholarships Council (No.202106470037).

\section*{Limitations}
Although we have tried to comprehensively analyze the literature of VideoQA research, we realize that we fail to cover and detail all the datasets and algorithms due to the thriving VideoQA research and the limited space. Hence, we complement the survey by maintaining a repository \url{https://github.com/VRU-NExT/VideoQA}. The repository contains the latest VideoQA papers, datasets, and their open-source implementations. We will periodly update the repository to trace the progress of the latest research. 

\clearpage

\bibliography{emnlp2022}
\bibliographystyle{acl_natbib}

\clearpage
\appendix

\setcounter{table}{0}
\setcounter{figure}{0}

\begin{table*}[h!]
\renewcommand{\thetable}{A\arabic{table}}
		\renewcommand\arraystretch{1.1}
		\center
		\caption{VideoQA datasets in the literature. (MTax: Modality-based Taxonomy, QTax: Question-based Taxonomy, Vid: VideoQA, MM: Multi-modal VideoQA, KB: Knowledge-based VideoQA, F: Factoid VideoQA, I: Inference VideoQA, Auto: automatic generation, Man: manual annotation, MC: multi-choice QA, OE: open-ended QA.)}
		\label{table:table_details_datasets}
		\scalebox{0.63}{
			\begin{tabular}{@{}llllllll@{}}
				\toprule\toprule
				Dataset & MTax & QTax &  Data Source &Goal
				& \#Video/\#QA & Annotation& Task \\ \midrule
				
				VideoQA(FiB)~\citep{zhu2017uncovering} & Vid & F&Multiple source & Temporal reasoning
				& 109K/390K & Auto & MC\\
				
				VideoQA~\citep{zeng2017leveraging} & Vid & F & Web videos & Description 
				& 18K/174K & Auto, Man & OE\\

				MSVD-QA~\citep{xu2017video}& Vid&F& Web videos & Description
				& 1.9K/50K &Auto & OE\\
				
				MSRVTT-QA~\citep{xu2017video}& Vid&  F& Web videos & Description 
				& 10K/243K &Auto & OE \\
				
				YouTube2Text-QA~\citep{zhao2017videoa} & Vid&  F & Web videos & Description 
				& 1.9K/48K & Auto & MC, OE\\
				
				MarioQA~\citep{mun2017marioqa} & Vid & F & Game & Temporal reasoning
				& 92K/92K & Auto & OE\\

				ActivityNet-QA~\citep{yu2019activitynet} & Vid &F & Web videos & Description 
				& 5.8K/58K & Man & OE\\
				
				EgoVQA~\citep{fan2019egovqa} & Vid & F & Egocentric videos & First-person VideoQA
				& 520/580 & Man & MC\\

				HowToVQA69M~\citep{yang2021just} & Vid&F&Web videos & Pre-training for downstream tasks
				& 69M/69M & Auto & OE\\

				iVQA~\citep{yang2021just} & Vid&F &Web videos& Removing language bias
				& 10K/10K  & Man & OE\\
				
				ASRL-QA~\citep{sadhu2021video} & Vid & F & Internet videos & VideoQA with phrases & 35K/162K & Auto & OE\\
				
				Charades-SRL-QA~\citep{sadhu2021video} & Vid & F & Crowd-Sourced & VideoQA with phrases & 9.5K/71K & Auto & OE\\
				
				WebVidVQA3M~\citep{yang2022learning} & Vid&F&Web videos & Pre-training for downstream tasks
				& 2M/3M & Auto & OE\\
				
				
				FIBER~\citep{castro2022fiber} & Vid&F&Web videos & Fill-in-the-blanks task with diverse answers
				& 28K/28K & Man & OE\\
				
				WildQA~\citep{castro2022wild} & Vid&F& In-the-wild videos & In-the-wild videos with evidence selection
				& 369/916 & Man & OE\\

				MovieQA~\citep{tapaswi2016movieqa} & MM&F& Movies & Text \& Visual story comprehension 
				& 6.7K/6.4K &Man & MC\\
				
				MovieFIB~\citep{maharaj2017dataset} & MM & F & Movies & Description 
				& 118K/348K& Auto & OE\\
				
				PororoQA~\citep{kim2017deepstory} & MM & F & Cartoon & Story comprehension 
				& 171/8.9K & Man & MC\\
				
				TVQA~\citep{lei2018tvqa} & MM & F & TV shows & Subtitle \& Concept comprehension 
				& 21K/152K &Man & MC\\
				
				TVQA+~\citep{lei2020tvqa+} & MM & F & TV shows & Spatio-temporal VideoQA
				& 4.1K/29K & Man & MC\\
				
				LifeQA~\citep{castro2020lifeqa} & MM & F  & Web videos & Real-life understanding 
				& 275/2.3K  & Man & MC\\
				
				How2QA~\citep{li2020hero}&MM & F & Web videos & Multimodal challenges 
				&22K/44K &Man &MC\\
				
				Env-QA~\citep{gao2021env}& MM & F & Egocentric videos & Exploring \& interacting with environments
				& 23K/85K & Auto, Man & OE\\
				
				Pano-AVQA~\citep{yun2021pano} & MM & F & 360$^{\circ}$ videos& Spherical spatial \& audio-visual relation
				& 5.4K/51.7K & Man & OE\\
				
				DramaQA~\citep{choi2021dramaqa} & MM & F & TV shows & Story comprehension
				& 23K/17K & Man & MC\\
				
				MUSIC-AVQA~\citep{li2022learning} & MM & F & Musical videos & Audio-Visual VideoQA
				& 9.3K/45K & Man & OE\\
				\hline
								
				TGIF-QA~\citep{jang2017tgif} & Vid & I& Animated GIF & Spatio-temporal reasoning
				& 71K/165K & Auto, Man & MC, OE\\
    
				SVQA~\citep{song2018explore} & Vid & I & Synthetic videos & Logical compositional questions
				& 12K/118K & Auto & OE\\
				
				Social-IQ~\citep{zadeh2019social} & MM&I& Web videos & Measuring social intelligence 
				& 1.2K/7.5K &Man & MC\\
				
				PsTuts-VQA~\citep{DBLP:conf/ijcai/ZhaoKXJ20} & KB & I  & Tutorial videos & Narrated instructional videos 
				& 76/17K & Man& MC\\
				
				KnowIT VQA~\citep{garcia2020knowit} & KB&I & TV shows & Knowledge in VideoQA
				& 12K/24K & Man & MC\\
				
				KnowIT-X VQA~\citep{wu2021transferring} & KB& I & TV shows & Transfer learning
				& 12K/21K & Man & MC\\
				
				NEWSKVQA~\citep{gupta2022newskvqa} & KB & I & News videos & Knowledge-based QA of news videos & 12K/1M & Auto & MC\\
												
				V2C-QA~\citep{fang2020video2commonsense} & Vid & I & Web videos & Commonsense reasoning
				& 1.5K/37K & Auto & OE\\
				
				TutorialVQA~\citep{colas2020tutorialvqa} & Vid & I & Tutorial videos & Multi-step \& non-factoid VideoQA
				& 408/6.1K & Man & OE \\
				
				CLEVRER~\citep{yi2020clevrer} & Vid&I& Synthetic videos & Temporal \& causal structures
				& 10K/305K& Auto & MC, OE \\
		
				TGIF-QA-R~\citep{peng2021progressive} & Vid & I & Animated GIF & Overcoming answer biases
				& 71K/165K & Auto & MC\\
				
				SUTD-TrafficQA~\citep{xu2021sutd} & Vid&I & Traffic scenes & Understanding \& inference in traffic
				& 10K/62K & Man & MC\\
				
				AGQA\citep{grunde2021agqa} & Vid & I& Homemade videos & Compositional reasoning 
				&9.6K/192M &Auto&OE \\ 
				
				AGQA 2.0\citep{gandhi2022measuring} & Vid & I & Homemade videos &
				Compositional consistency 
				&9.6K/4.55M&Auto&OE\\
				
				NExT-QA~\citep{xiao2021next} & Vid&I& Web videos & Causal \& temporal action interactions 
				& 5.4K/52K & Man & MC, OE\\
				
				STAR~\citep{wu2021star} & Vid &I & Homemade videos & Situated reasoning in real-world videos
				& 22K/60K & Auto & MC\\
                Causal-VidQA~\citep{li2022representation} & Vid &I & Web videos & Evidence \& commonsense reasoning
				& 26K/107K & Man & MC\\
                				
				VQuAD~\citep{gupta2022vquad} & Vid & I & Synthetic videos &
				Spatio \& temporal reasoning
				& 7K/1.3M& Auto & OE \\
				\bottomrule
		\end{tabular}}
	\end{table*}

\begin{sidewaysfigure}[h!]
\renewcommand{\thefigure}{A\arabic{figure}}
	\centering
	\includegraphics[height=0.26\textwidth, width=1.0\linewidth]{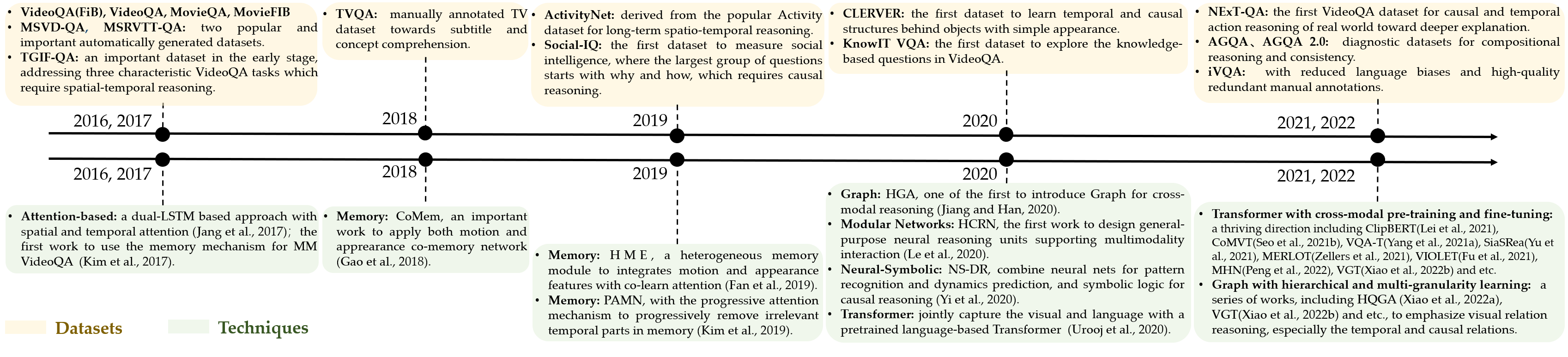}
	\caption{Timeline of VideoQA datesets and techniques. In the literature of VideoQA, datasets and techniques jointly evolve over time, and some of them influence each other.}
	\label{fig:app_history}
\end{sidewaysfigure}

\section{Appendix: Details of VideoQA datasets in the literature}
Due to limited space, details of VideoQA datasets are listed in Table~\ref{table:table_details_datasets}.

\section{Appendix: Timeline of VideoQA techniques}

In the literature, the VideoQA datasets and techniques jointly evolve in time (as shown in Figure~\ref{fig:app_history}). Some of the datasets and techniques influence each other. As the cross-modal pre-training and fine-tuning technique develops, the performance of early-stage datasets like TGIF-QA~\citep{jang2017tgif} and TVQA+~\citep{lei2020tvqa+} reaches unprecedentedly high (close to human performance, refer to Table~\ref{table:table_tc} and Table~\ref{table:table_mkn}). The new research focus turns to the more challenging VideoQA datasets like NExT-QA~\citep{xiao2021next}, which invokes complicated inference among multiple objects and relations. In turn, the inference QA datasets motivate new research interests in new techniques. CLEVRER~\citep{yi2020clevrer} has inspired new works using neuro-symbolic learning~\citep{yi2020clevrer,chen2021grounding,ding2021dynamic}, and NExT-QA has promoted a lot of recent works on graph models~\citep{xiao2021video,xiao2022video}. Diagnostic datasets like AGQA~\citep{grunde2021agqa} and AGQA 2.0~\citep{gandhi2022measuring} analyze existing methods by checking compositional consistency to examine whether they gain true understanding. These diagnostic datasets are promising to find the existing defects and motivate new methods~\citep{Qian2022DynamicSM}.

\end{document}